\documentclass{article}

\usepackage[preprint]{neurips_2026}

\usepackage[utf8]{inputenc}
\usepackage[T1]{fontenc}
\usepackage{hyperref}
\hypersetup{hidelinks}
\usepackage{xurl}
\usepackage{booktabs}
\usepackage{tabularx}
\usepackage{array}
\usepackage{placeins}
\usepackage[table]{xcolor}
\usepackage{amsmath}
\usepackage{amssymb}
\usepackage{microtype}
\usepackage{graphicx}
\usepackage{tikz}
\usetikzlibrary{arrows.meta}

\newcolumntype{Y}{>{\raggedright\arraybackslash}X}
\newcolumntype{C}[1]{>{\centering\arraybackslash}p{#1}}
\newcolumntype{L}[1]{>{\raggedright\arraybackslash}p{#1}}
\newcolumntype{M}[1]{>{\centering\arraybackslash}m{#1}}
\newcommand{\papertablestyle}{\normalsize\renewcommand{\arraystretch}{1.08}\setlength{\tabcolsep}{5pt}}
\newcommand{\papertablenote}[1]{\par\vspace{5pt}\begin{minipage}{\linewidth}\normalsize\raggedright #1\end{minipage}}
\newcommand{\methodA}{NL-SOP}
\newcommand{\methodB}{Prompt Graph}
\newcommand{\methodC}{MetaSkill}

\newcommand{\RuntimeTraceCount}{60}
\newcommand{\RuntimeExpectedNodes}{264}
\newcommand{\RuntimeCompletedNodes}{264}
\newcommand{\RuntimeDependencyEdges}{282}
\newcommand{\RuntimePrecedenceViolations}{0}
\newcommand{\RuntimeExplicitRefs}{270}
\newcommand{\RuntimePopulatedRefs}{270}
\newcommand{\RuntimeConcurrentTraces}{18}
\newcommand{\NumTasks}{20}
\newcommand{\NumRepeats}{3}
\newcommand{\ConformanceDependency}{\RuntimePrecedenceViolations/\RuntimeDependencyEdges}
\newcommand{\ConformanceBinding}{\RuntimePopulatedRefs/\RuntimeExplicitRefs}
\newcommand{\ConformanceParallel}{\RuntimeConcurrentTraces/\RuntimeTraceCount}

\newcommand{\AgentMatchedN}{60}
\newcommand{\AgentMatchedRepeats}{3}
\newcommand{\DeepCAWTL}{22/23/15}
\newcommand{\DeepCAPref}{0.558}
\newcommand{\DeepCACI}{[0.467, 0.650]}
\newcommand{\DeepCBWTL}{28/23/9}
\newcommand{\DeepCBPref}{0.658}
\newcommand{\DeepCBCI}{[0.575, 0.742]}
\newcommand{\AgentMatchedWTL}{26/13/21}
\newcommand{\AgentMatchedPref}{0.542}
\newcommand{\AgentMatchedCI}{[0.425, 0.650]}

\newcommand{\NativeAWTL}{23/22/15}
\newcommand{\NativeAPref}{0.567}
\newcommand{\NativeACI}{[0.433, 0.692]}
\newcommand{\NativeFWTL}{27/21/12}
\newcommand{\NativeFPref}{0.625}
\newcommand{\NativeFCI}{[0.525, 0.725]}
\newcommand{\NativeGWTL}{25/19/16}
\newcommand{\NativeGPref}{0.575}
\newcommand{\NativeGCI}{[0.433, 0.717]}

\newcommand{\GLMReviewedCAWTL}{23/22/15}
\newcommand{\GLMReviewedCAPref}{0.567}
\newcommand{\GLMReviewedCACI}{[0.433, 0.692]}
\newcommand{\GLMReviewedCBWTL}{22/22/16}
\newcommand{\GLMReviewedCBPref}{0.550}
\newcommand{\GLMReviewedCBCI}{[0.425, 0.667]}

\title{Where Natural Language Ends in Agents: Executable Skill Composition with MetaSkill}
\workshoptitle{Linguistic Principles for Foundation Models (LP4FM)}

\author{%
  \textbf{Meng Shu}$^{1,2}$ \quad
  \textbf{Wei He}$^{1}$ \quad
  \textbf{Hang Zhou}$^{1}$ \quad
  \textbf{Runke Liu}$^{1,3}$ \quad
  \textbf{Kai Han}$^{1,}$\thanks{Corresponding authors.} \quad
  \textbf{Yunhe Wang}$^{1,*}$ \\[5pt]
  \normalfont\small $^{1}$TokenRhythm Technologies \\
  \normalfont\small $^{2}$Xidian University \quad $^{3}$Beijing Institute of Technology \\[3pt]
  \normalfont\small\texttt{\href{mailto:23009100831@stu.xidian.edu.cn}{23009100831@stu.xidian.edu.cn},
  \href{mailto:3220242015@bit.edu.cn}{3220242015@bit.edu.cn},} \\
  \normalfont\small\texttt{\{\href{mailto:wei.he@tokenrhythm.ai}{wei.he}, \href{mailto:hang.zhou@tokenrhythm.ai}{hang.zhou}, \href{mailto:kai.han@tokenrhythm.ai}{kai.han}, \href{mailto:yunhe.wang@tokenrhythm.ai}{yunhe.wang}\}@tokenrhythm.ai}
}
\hypersetup{%
  pdftitle={Where Natural Language Ends in Agents: Executable Skill Composition with MetaSkill},
  pdfauthor={Meng Shu, Wei He, Hang Zhou, Runke Liu, Kai Han, Yunhe Wang},
  pdfsubject={MetaSkill: executable composition of Agent Skills},
  pdfkeywords={MetaSkill, Agent Skills, executable coordination, skill composition}
}

\makeatletter
\renewcommand{\@noticestring}{Outstanding Paper at the NeurIPS 2026 Workshop on Linguistic Principles for Foundation Models (LP4FM).}
\makeatother

\begin{document}

\maketitle

\begin{abstract}
Agent Skills express open-ended task knowledge in natural language, but composing them requires persistent execution order and upstream-output handoffs.
We propose \emph{selective formalization at Skill boundaries}: coordination relations with stable references and machine-checkable constraints become executable, while substantive task judgments remain within Skills.
We introduce \emph{\mbox{MetaSkill}}, a hybrid representation that preserves complete Skill packages and local instructions while making invocation dependencies and predecessor-text bindings executable through a directed acyclic graph (DAG) runtime.
This interface supports explicit coordination without prescribing how each Skill performs its local task.
Across 20 multi-Skill artifact tasks and two LLMs, we compare natural-language procedures, prompt-visible graphs, and \mbox{MetaSkill} derived from a common frozen plan per task, using the same designated Skill packages.
Additional comparisons evaluate \mbox{AgentSkillOS} with the same available Skill packages as \mbox{MetaSkill}, as well as agent systems using their configured tools.
Blinded artifact judgments yield favorable preference point estimates across all reported comparisons, with 95\% task-bootstrap intervals excluding parity against Prompt~Graph under \mbox{DeepSeek-V4-Flash} and \mbox{OpenClaw} under \mbox{GLM-5.2}.
In audited GLM-5.2 execution traces, all planned nodes complete and all declared precedence constraints are satisfied.
Together, these findings support \mbox{MetaSkill} as a practical interface for composing open-ended Skills, combining local linguistic flexibility with executable dependencies, explicit predecessor-text bindings, and auditable coordination.
Code is available at: \href{https://github.com/TokenRhythm/opensquilla/tree/paper/metaskill-lp4fm-2026}{{\urlstyle{same}\nolinkurl{github.com/TokenRhythm/opensquilla}}}.
\end{abstract}

\section{Introduction}

Agent Skills package reusable instructions, scripts, templates, and domain resources without reducing open-ended work to a fixed function signature \citep{agent_skills_specification}.  This flexibility supports artifact-rich tasks such as analysis, document production, and presentation design.  When several Skills cooperate, however, the agent must maintain ordering, independent branches, joins, and artifact handoffs across invocations.  Individually plausible steps can still reuse the wrong output or start a dependent step too early.

A natural-language standard operating procedure (NL-SOP) expresses these constraints readably but leaves their preservation to the model.  Workflow and agent languages make control flow and state executable \citep{agentspex,flowagent}, while grammars, schemas, and hybrid language--symbolic programs show how representational form can shape generation and reasoning \citep{grammar_prompting,binder,shaping_schema}.  For open-ended Skills, the design question is where the boundary between language and executable structure should lie.

\begin{figure}[t]
\centering
\begingroup
\definecolor{MSInk}{HTML}{252D36}
\definecolor{MSGray}{HTML}{626C78}
\definecolor{MSRule}{HTML}{CBD2DA}
\definecolor{MSBlue}{HTML}{315F91}
\definecolor{MSTeal}{HTML}{217D73}
\definecolor{MSOrange}{HTML}{9B681F}
\definecolor{MSBlueLight}{HTML}{F0F5FA}
\definecolor{MSTealLight}{HTML}{EFF7F5}
\definecolor{MSOrangeLight}{HTML}{FBF5EB}
\definecolor{MSGrayLight}{HTML}{F3F5F7}
\begin{tikzpicture}[
  x=1mm,y=-1mm,
  every node/.style={inner sep=0pt,outer sep=0pt,text=MSInk},
  title/.style={font=\sffamily\bfseries\fontsize{9.2}{10}\selectfont,anchor=west},
  body/.style={font=\sffamily\fontsize{8}{9.2}\selectfont,anchor=west},
  note/.style={font=\sffamily\fontsize{8}{9.2}\selectfont,anchor=west},
  stephead/.style={font=\sffamily\bfseries\fontsize{8}{9.2}\selectfont,anchor=west},
  dep/.style={-{Latex[length=1.65mm,width=1.15mm]},draw=MSBlue,line width=.65pt},
  binding/.style={-{Latex[length=1.65mm,width=1.15mm]},draw=MSTeal,line width=.6pt,dash pattern=on 2.1pt off 1.6pt}
]
\path[use as bounding box] (0,0) rectangle (139.7,63.3);

\node[title] at (1,3.2) {(a)\enspace MetaSkill representation};
\node[note] at (1,8.3)
  {\textcolor{MSOrange}{Language within Skills}; \textcolor{MSBlue}{executable coordination across Skills}};
\node[body] at (1,13.2) {$P_i=(\Gamma_i,L_i)$};
\draw[dep] (31,13.2)--(37,13.2);
\node[note,text=MSBlue] at (38.5,13.2) {dependency};
\draw[binding] (62,13.2)--(68,13.2);
\node[note,text=MSTeal] at (69.5,13.2) {text binding};

\newcommand{\msnode}[4]{%
  \draw[fill=white,draw=MSRule,line width=.45pt,rounded corners=.5pt]
    (#1,#2) rectangle ({#1+23},{#2+#4});
  \draw[MSBlue,line width=1.15pt] (#1,#2)--(#1,{#2+#4});
  \node[body,text=MSBlue] at ({#1+1.25},{#2+2.65})
    {$v_{#3}:\,\kappa(v_{#3})=S_{#3}$};
  \draw[MSRule,line width=.35pt] ({#1+1.25},{#2+4.95})--({#1+21.75},{#2+4.95});
  \node[stephead] at ({#1+1.25},{#2+7.9}) {complete Skill};
  \node[body,text=MSOrange] at ({#1+1.25},{#2+11.55}) {instruction $\ell_{i#3}$};
}
\msnode{.75}{26}{1}{13.8}
\msnode{30}{18.5}{2}{13.8}
\msnode{30}{36.8}{3}{13.8}
\msnode{60}{24.9}{4}{17.3}
\node[body,text=MSBlue] at (61.25,40.1)
  {after $v_2$ and $v_3$};

\draw[dep] (23.75,32.9)--(30,25.4);
\draw[dep] (23.75,32.9)--(30,43.7);
\draw[dep] (53,25.4)--(60,33.55);
\draw[dep] (53,43.7)--(60,33.55);

\node[note,text=MSTeal,anchor=center] at (42,52.9) {render into downstream argument};
\node[body,text=MSTeal] at (1,55.9) {\texttt{outputs[v2]}};
\draw[binding] (23,55.9)--(60,55.9);
\node[body,text=MSTeal,anchor=east] at (83,55.9) {argument of $v_4$};
\draw[MSRule,line width=.4pt] (1,58.2)--(83,58.2);
\node[note,text=MSGray] at (1,60.9) {Files remain in a shared task workspace};

\draw[MSRule,line width=.45pt] (86,1.1)--(86,62.1);
\node[title] at (88.4,3.2) {(b)\enspace Runtime execution};
\node[stephead] at (88.4,8.7) {Parse and validate};
\node[note,text=MSGray,align=left] at (88.4,13.6)
  {Node IDs; dependency references\\acyclicity};
\draw[MSRule,line width=.4pt] (88.4,17.1)--(136,17.1);
\node[note,text=MSGray] at (88.4,19.2) {Scheduling loop};

\draw[MSRule,line width=.6pt] (90.8,24.9)--(90.8,55.5);
\newcommand{\msstep}[5]{%
  \draw[draw=none,fill=#3,rounded corners=1.5pt]
    (94.1,#1) rectangle (136,{#1+7.8});
  \node[circle,draw=#4,line width=.6pt,fill=white,
    minimum size=4.1mm,font=\sffamily\bfseries\fontsize{8}{9}\selectfont,
    text=#4] at (90.8,{#1+3.9}) {#2};
  \node[stephead] at (95.8,{#1+2.25}) {#5};
}
\msstep{21}{1}{MSBlueLight}{MSBlue}{Compute ready set}
\node[body,text=MSGray] at (95.8,26.55)
  {$Q_t=\{v\in U_t:\operatorname{pred}(v)\subseteq T_t\}$};
\msstep{31.2}{2}{MSTealLight}{MSTeal}{Render argument bindings}
\node[note,text=MSGray] at (95.8,36.75)
  {predecessor text $\rightarrow$ arguments};
\msstep{41.4}{3}{MSOrangeLight}{MSOrange}{Dispatch ready subset}
\node[body,text=MSGray] at (95.8,46.95)
  {$F_t\subseteq Q_t,\quad |F_t|\leq K$};
\msstep{51.6}{4}{MSGrayLight}{MSGray}{Store node outputs}
\node[note,text=MSGray] at (95.8,57.15)
  {store text; release dependents};

\draw[-{Latex[length=1.4mm,width=1.0mm]},draw=MSGray,line width=.5pt,
  dash pattern=on 2pt off 1.5pt,rounded corners=1mm]
  (136.1,55.5)--(138.9,55.5)--(138.9,24.9)--(136.1,24.9);
\end{tikzpicture}%
\endgroup
\caption{\textbf{MetaSkill selectively formalizes coordination across Skill invocations.}
(a) Nodes invoke complete Skill packages via $\kappa_i$, with local instructions in $L_i$ and executable coordination $\Gamma_i$. The binding supplies $v_2$'s text to $v_4$; files remain in a shared workspace. (b) After one structural validation, each runtime scheduling step computes readiness, renders bindings, dispatches at most $K$ ready nodes, and stores outputs to release dependents.}
\label{fig:overview}
\end{figure}
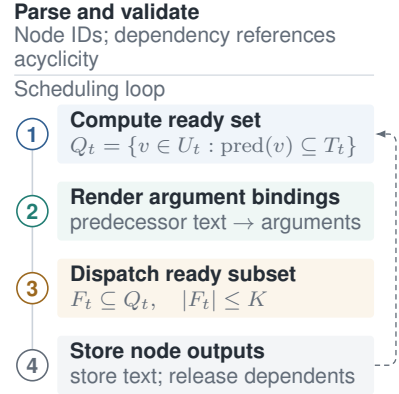

We place that boundary between Skill invocations.  MetaSkill preserves each complete Skill package and the local instructions for each invocation while expressing dependencies and predecessor-text bindings in a DAG executed by the runtime.  The runtime validates the graph, dispatches dependency-ready nodes, renders named predecessor text into downstream arguments, and records outputs; generated files remain in a shared workspace.  Figure~\ref{fig:overview} separates semantic freedom within Skills from persistent coordination across them.

We test this boundary using \NumTasks{} multi-Skill AgentSkillOS benchmark tasks \citep{agentskillos}.  A controlled two-backbone study compares three coordination interfaces derived from each frozen plan: \methodA{}, \methodB{}, and \methodC{}.  External comparisons cover AgentSkillOS with matched Skill availability, plus OpenSquilla, OpenClaw, and Hermes Agent using their configured tool stacks.  Artifact-preference point estimates favor MetaSkill across all reported comparisons. The 95\% task-bootstrap intervals exclude parity against \methodB{} under DeepSeek-V4-Flash and OpenClaw under GLM-5.2.

Our contributions are threefold. First, we formulate \emph{selective formalization at Skill boundaries}, with criteria for identifying persistent, referentially stable, and machine-observable coordination commitments. Second, MetaSkill gives these commitments executable semantics: invocation dependencies govern dispatch, and predecessor-text bindings construct downstream arguments, while complete Skill packages retain their local task instructions and resources. Third, we evaluate this interface through comparisons derived from common frozen plans, matched-Skills and agent-stack comparisons, and trace auditing, providing evidence about both artifact quality and declared execution properties.

\section{Related work}

\paragraph{Tool-using agents and Skill ecosystems.}
ReAct interleaves reasoning and actions \citep{react}; Voyager accumulates executable programs in a retrievable Skill library for embodied lifelong learning \citep{voyager}.  AgentSkillOS couples Skill retrieval, generated DAG planning, execution, and an artifact-rich benchmark \citep{agentskillos}.  AgentSkillOS is our closest system and task precedent. Its comparison of generated DAG execution with native flat Skill invocation motivates studying composition. Our controlled study fixes designated Skills and one shared frozen source plan per task to compare handoff representation and execution; our system comparison under matched Skill availability also includes planning.  The OpenSquilla technical report describes a Meta-Skill orchestration layer \citep{opensquilla2026}. The present study examines selective formalization of cross-Skill dependencies and artifact handoffs through a minimal coordination representation, runtime semantics, and empirical evaluation.

\paragraph{Agent and workflow languages.}
DSPy represents LM pipelines as declarative modules and optimizes prompts or demonstrations \citep{dspy}; LLMCompiler schedules dependency-aware function-call plans \citep{llmcompiler}.  AgentSPEX provides typed steps, branches, loops, explicit state, and runtime services \citep{agentspex}; FlowAgent combines natural-language procedures, code, and controller supervision for workflow compliance \citep{flowagent}.  MetaSkill composes complete heterogeneous Skill packages using a compact acyclic coordination core and predecessor-text bindings, without rewriting Skill internals into a workflow-specific language. The distinction concerns composition units: function-call graphs coordinate individual operations, whereas MetaSkill nodes retain ordinary Skills' instructions and supporting resources.

\paragraph{Coordination and computation.}
Coordination languages separate computation within activities from mechanisms that connect and synchronize them \citep{gelernter1992coordination}. MetaSkill applies this separation at a complete Skill invocation: the package retains its instructions and resources, while the composition declares dependencies and predecessor-text bindings for runtime scheduling and argument construction. These commitments can be checked through invocation-level records without prescribing artifact construction.

\paragraph{Hybrid representations.}
Grammar Prompting supplies grammars for domain-specific generation \citep{grammar_prompting}; BINDER embeds LM calls inside symbolic programs \citep{binder}; shaping schema studies how representation can organize and activate model knowledge \citep{shaping_schema}. PDL places prompts within executable YAML blocks that contribute results and context to later program blocks, and provides templates, functions, and file inclusion for composing programs \citep{vaziri2024pdl}. MetaSkill focuses on relations among invocations of complete Skill packages. A declared predecessor-text binding selects the source of upstream text for a downstream argument; the receiving Skill still interprets that context and decides how to use available files. Dependencies govern runtime dispatch, and predecessor-text bindings govern argument construction, making these coordination relations inspectable at invocation boundaries.

\paragraph{Workflow benchmarks.}
FlowBench studies workflow knowledge formats \citep{flowbench}, TaskBench evaluates decomposition, tool selection, and parameter prediction \citep{taskbench}, WorkflowLLM builds an orchestration-generation corpus \citep{workflowllm}, and WorFBench measures generated sequence and graph fidelity \citep{worfbench}.  AgentSkillOS tasks supply multiple complete Skills and inspectable deliverables. Whereas generation benchmarks ask whether a model can recover a workflow, our study fixes one source plan per task and evaluates artifacts produced using prose, prompt-visible fields, or executable structure. Holding plan synthesis constant focuses the comparison on how plan information is presented and used during execution.

\section{Where natural language ends in agents}
\label{sec:method}

\subsection{Representation boundary}

Each MetaSkill node invokes a complete ordinary Skill, including its instructions, scripts, references, templates, and assets, with a local natural-language instruction.  Choices such as evidence synthesis, page layout, and prose revision remain language-driven.  The composition layer adds four coordination elements: stable node identity, package selection, predecessor dependencies, and predecessor-text bindings.

\paragraph{A criterion for crossing the boundary.}
We use three questions to decide whether a relation belongs in executable structure.  First, is it \emph{persistent}: must the same commitment survive across separately invoked Skills rather than be reconsidered inside one Skill?  Second, is it \emph{referentially stable}: does it name a specific node, predecessor, or stored output rather than an open-ended semantic preference?  Third, is it \emph{machine-observable}: can a runtime determine whether the relation was followed without judging the substantive quality of an artifact?  Predecessor completion and the identity of supplied upstream text satisfy all three criteria.  Instructions such as \emph{use a persuasive narrative} or \emph{choose the strongest evidence} leave interpretation to the invoked Skill and model.  The graph records durable relational commitments; natural language continues to represent the search and judgment that make heterogeneous Skills useful and remain open to interpretation within each Skill.  This selective boundary makes coordination enforceable without treating the entire task as a closed formal program.

For task instance $i$, the hybrid plan is
\begin{equation}
P_i=(\Gamma_i,L_i),\qquad
\Gamma_i=(V_i,E_i,\kappa_i,b_i),\qquad
L_i=\{\ell_{iv}\}_{v\in V_i}.
\label{eq:representation}
\end{equation}
Here $V_i$ is the invocation-node set, $E_i\subseteq V_i\times V_i$ an acyclic dependency relation, $\kappa_i(v)$ a Skill package, and $\ell_{iv}$ its local instruction.  The binding map $b_i$ supplies argument templates referencing task input or stored predecessor text.  Thus $\Gamma_i$ provides executable relations while open semantics remain in $L_i$ and the complete packages.

The unit of composition is a Skill invocation, not a tool call or typed data-processing operator.  The same package can participate in different MetaSkills without rewriting its internals; a task requiring only one Skill invocation needs no composition layer.  This minimal coordination layer preserves heterogeneous resources while giving the runtime the structure needed to enforce precedence and identify declared handoffs.  Package and coordination reuse are separate: a Skill can appear in different compositions, each invocation with its own objective and bindings.  Moving a relation into the graph fixes the commitment that must survive those invocations, without prescribing how a Skill synthesizes evidence or constructs an artifact.

\subsection{Runtime execution}

The runtime gives the representation operational force.  Before execution, it parses the composition and rejects duplicate node identifiers, undefined dependencies, and dependency cycles.  Let $T_t$ be the completed nodes after transition $t$ and $U_t$ the unstarted nodes.  The ready set is
\begin{equation}
Q_t=\{v\in U_t:\operatorname{pred}(v)\subseteq T_t\},
\qquad F_t\subseteq Q_t,\qquad |F_t|\le K,
\label{eq:ready}
\end{equation}
where $F_t$ is the subset dispatched in the next wave.  Independent ready nodes may execute concurrently; a fan-in node becomes eligible only after all predecessors complete.  Readiness permits parallelism without requiring every independent branch to overlap in wall-clock time.

At dispatch, Jinja templates in a node's argument dictionary are rendered against task input and the mapping of completed-node text outputs.  In the experimental MetaSkills, each inserted predecessor text is capped at 6000 characters; generated files remain available independently in the shared workspace.  The runtime then invokes the complete Skill package in host harness $H$ (the agent execution framework):
\begin{equation}
a_{iv}=\operatorname{Render}\!\left(b_i(v);x_i,O_t\right),\qquad
o_{iv}=\operatorname{Exec}_{H}\!\left(\kappa_i(v),\ell_{iv},a_{iv};\xi_{iv}\right).
\label{eq:execution}
\end{equation}
Each invocation uses a fresh agent context containing Skill instructions, original task $x_i$, $\ell_{iv}$, and $a_{iv}$. The runtime stores the invocation's accumulated visible assistant text as $o_{iv}$ by node identifier and releases newly ready dependents.  Here $O_t$ maps completed nodes to text outputs and $\xi_{iv}$ captures model and environment nondeterminism.  Dependencies control \emph{eligibility}; bindings construct \emph{context}.  Files remain workspace-mediated rather than serialized through typed ports.

\paragraph{A running composition example.}
In Figure~\ref{fig:overview}, completing $v_1$ makes $v_2$ and $v_3$ eligible; $v_4$ waits for both. Dependencies govern when invocations may start; the illustrated binding supplies $v_2$'s stored text to an argument of $v_4$. Although $v_3$ is a prerequisite, its completion alone does not supply its text to this argument. Files from both branches remain accessible in the shared workspace. The example distinguishes execution order, the identity of supplied context, and artifact access.

Under \methodA{}, the agent interprets the prerequisite statements and infers the text handoff.  \methodB{} supplies the named reference explicitly but leaves the agent responsible for realizing it during execution.  \methodC{} makes readiness and text insertion runtime operations, while each invoked Skill still decides how to use its context and files.  

\paragraph{Trace conformance.}For a trace $\tau$ recording one completed invocation per planned node, let $t_{\mathrm{start}}^\tau(v)$ and $t_{\mathrm{finish}}^\tau(v)$ denote the recorded start and completion times of node $v$.  We define precedence conformance as
\begin{equation}
\operatorname{PrecedenceOK}(\tau,E_i)
\iff
\forall (u,v)\in E_i,\quad
 t_{\mathrm{finish}}^\tau(u)\leq t_{\mathrm{start}}^\tau(v).
\label{eq:precedence-conformance}
\end{equation}
This condition imposes no additional order on nodes unrelated by a dependency path.  For each declared predecessor-text reference, we separately check that the referenced node exists in the trace and that the corresponding rendered input value is non-empty.  These checks concern recorded coordination; artifact quality is evaluated separately.

\subsection{Plan acquisition and scope}

A MetaSkill may be authored directly, adapted from an SOP, or proposed by an upstream plan generator.  The present study evaluates execution of a supplied plan.  For each task, we prepared one plan from its specification and designated Skills, checked its structure, and froze it before the reported evaluation.  A manifest records provenance, hashes, and rendering paths; the evaluation protocol keeps the frozen plan fixed, precluding revision based on evaluator outputs or execution history.  All three conditions (\methodA{}, \methodB{}, and \methodC{}) derive their inputs from this frozen plan; an input renderer selects and formats the plan fields supplied to a condition.

\section{Experimental design}
\label{sec:experiment}

We ask three empirical questions and one mechanism check.  \textbf{RQ1:} Under inputs derived from a shared frozen source plan, how does MetaSkill compare with NL-SOP and Prompt Graph in blinded artifact quality? \textbf{RQ2:} How does MetaSkill compare with AgentSkillOS in artifact quality when both systems receive the same available Skills?  \textbf{RQ3:} When OpenSquilla, OpenClaw, and Hermes Agent execute the frozen NL-SOP input, how do their artifacts compare directly with MetaSkill?  The mechanism check asks whether the MetaSkill runtime executes frozen dependencies and bindings as declared.

\subsection{Tasks, Skills, and evaluation protocol}

We use all \NumTasks{} AgentSkillOS benchmark tasks whose released definitions designate two to four Skills, excluding the 10 one-Skill tasks because our hypothesis concerns cross-Skill relations \citep{agentskillos}.  The set spans five task categories and yields DOCX, PPTX, HTML, JSON, image, and video artifacts.  Within the controlled study, every condition receives the same complete designated Skill packages, including non-markdown resources.  The frozen plans include both fan-in and nodes that can become ready together, so the representation exercises the relations studied here.

For each task and coordination interface, we perform \NumRepeats{} executions per backbone using \texttt{z-ai/glm-5.2} \citep{glm5team2026glm5vibecodingagentic} and DeepSeek-V4-Flash \citep{deepseekai2026deepseekv4}. Each backbone evaluation contains 180 runs and 360 judge calls covering both display orders of each artifact pair.  The plan and input renderer remain fixed across repetitions; model sampling and tool execution may vary.  The task is the statistical unit.  The AgentSkillOS matched-Skills and agent-stack comparisons each cover \AgentMatchedRepeats{} executions per task (\AgentMatchedN{} task--repetition pairs).

\subsection{Inputs derived from a shared frozen plan}

The coordination interfaces are deterministic projections of one shared frozen source plan:
\begin{equation}
\begin{aligned}
\text{NL-SOP} &= (\phi_{\mathrm{NL}}(P_i),\operatorname{Agent}_{H}),\\
\text{Prompt Graph} &= (\phi_{\mathrm{G}}(P_i),\operatorname{Agent}_{H}),\\
\text{MetaSkill} &= (\phi_{\mathrm{G}}(P_i),\operatorname{Runtime}_{H}).
\end{aligned}
\label{eq:intervention}
\end{equation}
$\phi_{\mathrm{NL}}$ renders node identity, Skill, objective, prerequisites, and expected result as a readable stepwise SOP; it does not include the frozen source plan's fields that identify predecessor text supplied to downstream input arguments.  $\phi_{\mathrm{G}}$ exposes the full records, including named input references.  In \methodB{}, the host agent sees those fields and remains responsible for maintaining execution state; in \methodC{}, the MetaSkill runtime supplies predecessor text in the invocation inputs and executes the graph.  All conditions run in the same host harness with the same backbone and complete Skill packages.

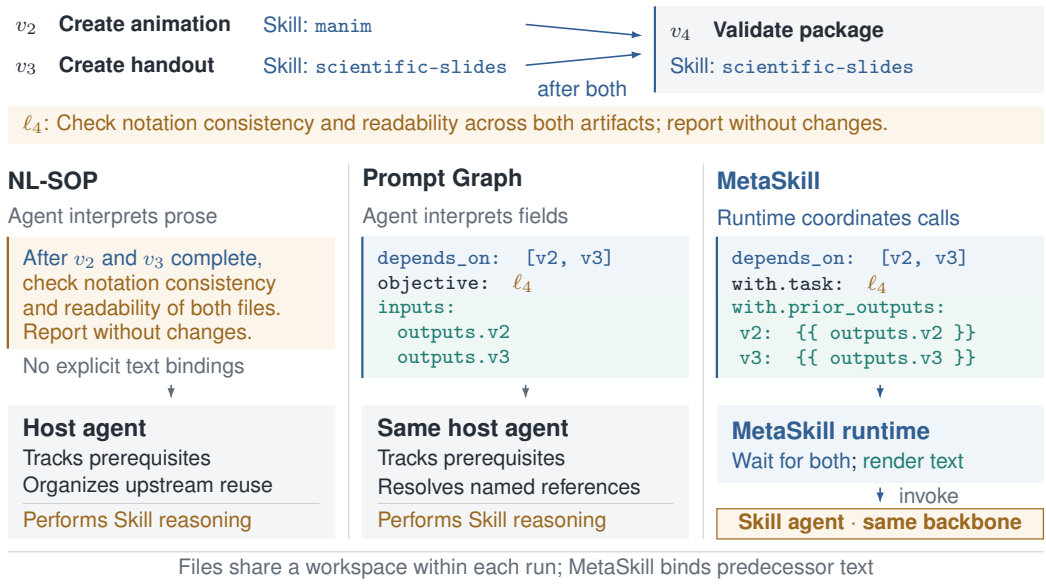
\begin{figure}[!htbp]
\centering
\begingroup
\definecolor{ABCInk}{HTML}{252D36}
\definecolor{ABCGray}{HTML}{626C78}
\definecolor{ABCRule}{HTML}{CBD2DA}
\definecolor{ABCBlue}{HTML}{315F91}
\definecolor{ABCTeal}{HTML}{217D73}
\definecolor{ABCAmber}{HTML}{9B681F}
\definecolor{ABCBlueLight}{HTML}{F0F5FA}
\definecolor{ABCTealLight}{HTML}{EFF7F5}
\definecolor{ABCAmberLight}{HTML}{FBF5EB}
\definecolor{ABCGrayLight}{HTML}{F3F5F7}
\begin{tikzpicture}[
  x=1mm,y=-1mm,
  every node/.style={inner sep=0pt,outer sep=0pt,text=ABCInk},
  abcbody/.style={font=\sffamily\fontsize{8}{9.3}\selectfont,anchor=west},
  abccode/.style={font=\ttfamily\fontsize{8}{9.3}\selectfont,anchor=west},
  abchead/.style={font=\sffamily\bfseries\fontsize{9}{10.4}\selectfont,anchor=west},
  abcstep/.style={font=\sffamily\bfseries\fontsize{8}{9.3}\selectfont,anchor=west},
  abcarrow/.style={-{Latex[length=1.45mm,width=1mm]},line width=.6pt,draw=ABCGray}
]
\path[use as bounding box] (0,0) rectangle (139.7,83.5);

\node[abchead] at (1.3,3) {Shared task: derivative teaching package};
\node[abcbody,text=ABCGray,anchor=east] at (138.3,3)
  {Animation (MP4) + handout ($\geq 8$ pages)};
\draw[ABCRule,line width=.4pt] (1.3,6.1)--(138.3,6.1);
\node[abcstep] at (2.3,9.6) {$v_2$\quad Create animation};
\node[abcbody,text=ABCBlue] at (35.1,9.6) {Skill: \texttt{manim}};
\node[abcstep] at (2.3,15) {$v_3$\quad Create handout};
\node[abcbody,text=ABCBlue] at (35.1,15) {Skill: \texttt{scientific-slides}};
\draw[abcarrow,draw=ABCBlue] (69.8,9.6)--(85.3,11.1);
\draw[abcarrow,draw=ABCBlue] (69.8,15)--(85.3,13.5);
\node[abcbody,text=ABCBlue,anchor=center] at (77.3,18) {after both};
\fill[ABCGrayLight] (86.9,7.2) rectangle (138.3,18.6);
\draw[ABCBlue,line width=.8pt] (86.9,7.2)--(86.9,18.6);
\node[abcstep] at (88.9,10.6) {$v_4$\quad Validate package};
\node[abcbody,text=ABCBlue] at (88.9,15) {Skill: \texttt{scientific-slides}};

\fill[ABCAmberLight] (1.3,20.4) rectangle (138.3,25.3);
\node[abcbody,text=ABCAmber] at (3.3,22.85)
  {$\ell_4$: Check notation consistency and readability across both artifacts; report without changes.};

\foreach \xx in {46.4,93.3}{\draw[ABCRule,line width=.4pt] (\xx,28)--(\xx,77.7);}
\node[abchead] at (1.3,30.2) {NL-SOP};
\node[abchead] at (48.2,30.2) {Prompt Graph};
\node[abchead,text=ABCBlue] at (95.1,30.2) {MetaSkill};
\node[abcbody,text=ABCGray] at (1.3,35.2) {Agent interprets prose};
\node[abcbody,text=ABCGray] at (48.2,35.2) {Agent interprets fields};
\node[abcbody,text=ABCBlue] at (95.1,35.2) {Runtime coordinates calls};

\fill[ABCAmberLight] (1.3,37.7) rectangle (44.5,52.5);
\draw[ABCAmber,line width=.85pt] (1.3,37.7)--(1.3,52.5);
\node[abcbody,text=ABCBlue] at (3.3,40.7) {After $v_2$ and $v_3$ complete,};
\node[abcbody,text=ABCAmber] at (3.3,44) {check notation consistency};
\node[abcbody,text=ABCAmber] at (3.3,47.3) {and readability of both files.};
\node[abcbody,text=ABCAmber] at (3.3,50.6) {Report without changes.};
\node[abcbody,text=ABCGray] at (3.3,55.1) {No explicit text bindings};

\fill[ABCBlueLight] (48.2,37.7) rectangle (91.4,56.4);
\fill[ABCTealLight] (48.2,45.6) rectangle (91.4,56.4);
\draw[ABCBlue,line width=.85pt] (48.2,37.7)--(48.2,56.4);
\node[abccode,text=ABCBlue] at (50.2,40.7) {depends\_on: [v2, v3]};
\node[abccode] at (50.2,44) {objective: \textcolor{ABCAmber}{$\ell_4$}};
\node[abccode,text=ABCTeal] at (50.2,47.3) {inputs:};
\node[abccode,text=ABCTeal] at (52.8,50.6) {outputs.v2};
\node[abccode,text=ABCTeal] at (52.8,53.9) {outputs.v3};

\fill[ABCBlueLight] (95.1,37.7) rectangle (138.3,56.4);
\fill[ABCTealLight] (95.1,45.6) rectangle (138.3,56.4);
\draw[ABCBlue,line width=.85pt] (95.1,37.7)--(95.1,56.4);
\node[abccode,text=ABCBlue] at (97.1,40.7) {depends\_on: [v2, v3]};
\node[abccode] at (97.1,44) {with.task: \textcolor{ABCAmber}{$\ell_4$}};
\node[abccode,text=ABCTeal] at (97.1,47.3) {with.prior\_outputs:};
\node[abccode,text=ABCTeal] at (98.1,50.6) {v2: \{\{ outputs.v2 \}\}};
\node[abccode,text=ABCTeal] at (98.1,53.9) {v3: \{\{ outputs.v3 \}\}};

\draw[abcarrow] (22.9,57.3)--(22.9,59.2);
\draw[abcarrow] (69.8,57.3)--(69.8,59.2);
\draw[abcarrow,draw=ABCBlue] (116.7,57.3)--(116.7,59.2);

\fill[ABCGrayLight] (1.3,60) rectangle (44.5,77.7);
\fill[ABCGrayLight] (48.2,60) rectangle (91.4,77.7);
\node[abchead] at (3.3,63.3) {Host agent};
\node[abchead] at (50.2,63.3) {Same host agent};
\node[abcbody] at (3.3,67.2) {Tracks prerequisites};
\node[abcbody] at (3.3,70.7) {Organizes upstream reuse};
\node[abcbody] at (50.2,67.2) {Tracks prerequisites};
\node[abcbody] at (50.2,70.7) {Resolves named references};
\draw[ABCRule,line width=.35pt] (3.3,73)--(42.5,73);
\draw[ABCRule,line width=.35pt] (50.2,73)--(89.4,73);
\node[abcbody,text=ABCAmber] at (3.3,75.4) {Performs Skill reasoning};
\node[abcbody,text=ABCAmber] at (50.2,75.4) {Performs Skill reasoning};

\fill[ABCBlueLight] (95.1,60) rectangle (138.3,70.3);
\node[abchead,text=ABCBlue] at (97.1,63.3) {MetaSkill runtime};
\node[abcbody] at (97.1,67.5)
  {\textcolor{ABCBlue}{Wait for both}; \textcolor{ABCTeal}{render text}};
\draw[abcarrow,draw=ABCBlue] (116.7,70.9)--(116.7,73);
\node[abcbody,text=ABCGray] at (119.2,71.9) {invoke};
\draw[fill=ABCAmberLight,draw=ABCAmber,line width=.45pt]
  (95.1,73.7) rectangle (138.3,77.7);
\node[abcstep,text=ABCAmber,anchor=center] at (116.7,75.7)
  {Skill agent $\cdot$ same backbone};

\draw[ABCRule,line width=.4pt] (1.3,79.4)--(138.3,79.4);
\node[abcbody,text=ABCGray,anchor=center] at (69.85,81.6)
  {Files share a workspace within each run; MetaSkill binds predecessor text};
\end{tikzpicture}%
\endgroup
\caption{\textbf{One task, three coordination interfaces.}
NL-SOP and Prompt Graph coordinate within the host agent; MetaSkill separates runtime coordination from model-driven Skill execution. All share the frozen task, complete Skills, objectives, expected outputs, backbone, and host harness. Node IDs are case-local; canonical Skill names and abridged interfaces are shown. The illustration omits the shared upstream planning node and text-truncation filters. MetaSkill's \texttt{depends\_on} lists prerequisites, \texttt{with} contains invocation arguments, and \texttt{outputs.v2} and \texttt{outputs.v3} reference stored predecessor text.}
\label{fig:interfaces-case}
\end{figure}

Figure~\ref{fig:interfaces-case} instantiates these interfaces on a frozen derivative-teaching task.    \methodB{} versus \methodA{} adds explicit handoff identity as well as field structure.  \methodC{} versus \methodB{} retains graph content but changes the executor, context construction, dispatch, and error handling together, evaluating these runtime components as a combined system.  \methodC{} versus \methodA{} compares the complete method.  All input renderers share the request, Skills, objectives, precedence, and expected outputs, but only \methodB{} and \methodC{} expose named input references.

\subsection{Comparison scope and pairing}
\label{sec:estimands}

Each comparison pairs artifacts by task and repetition. The reported GLM-5.2 controlled, matched-Skills, and agent-stack comparisons share the same MetaSkill artifact set.

The AgentSkillOS comparison fixes Skill availability: each AgentSkillOS run receives exactly the Skill set used by its paired \methodC{} run but retains AgentSkillOS planning, execution, context management, and error handling. The quality analysis covers \AgentMatchedN{} task--repetition pairs and uses the same blinded judging procedure for both systems. This comparison evaluates planning and execution given the available Skills; it does not evaluate Skill retrieval.

For the agent-stack comparison, OpenSquilla, OpenClaw \citep{openclaw_software}, and Hermes Agent \citep{hermes_agent_software} each receive the frozen NL-SOP input under GLM-5.2 and use their configured tool stacks.  OpenSquilla \citep{opensquilla2026} is the host agent for MetaSkill. The OpenSquilla (NL-SOP) baseline executes the frozen natural-language procedure without the MetaSkill runtime, whereas the MetaSkill condition uses runtime-managed dependencies and predecessor-text bindings.

\subsection{Evaluation measures}

The reported artifact-quality comparisons each cover 60 task--repetition pairs. We use blinded pairwise artifact preference as the main outcome; it measures relative judged quality rather than absolute task success or a global ranking.

\paragraph{Blinded artifact quality.}
We adapt AgentSkillOS's pairwise, order-swapped procedure \citep{agentskillos} for all three interface pairs and the external comparisons. Table~\ref{tab:abc-results} reports the two interface comparisons that directly involve MetaSkill on each backbone.  Evaluator-referenced artifacts receive opaque aliases; the judge never sees the method mapping.  A fixed \texttt{anthropic/claude-opus-4.5} judge \citep{claude_opus_45} at temperature zero evaluates both display orders, with conflicting valid winners merged to a tie.  Using AgentSkillOS's comparison prompt, the judge receives the task description and both systems' artifacts, assessing quality, completeness, correctness, and aesthetics. JSON and other text files are included directly (up to 5000 characters per file); visual artifacts are rendered as images, and videos as uniformly sampled frames with metadata. We send text and images through OpenRouter, with at most 30 images per side and the first frame used for GIFs.

For method $c$ against $d$, let $q_{ir}^{c>d}\in\{0,0.5,1\}$ encode loss, tie, or win for task $i$ and repetition $r$.  With $N=\NumTasks$ and $R=\NumRepeats$,
\begin{equation}
\widehat p_{c>d}=\frac{1}{NR}\sum_{i=1}^{N}\sum_{r=1}^{R}q_{ir}^{c>d}.
\label{eq:preference}
\end{equation}
We report W/T/L, half-tie preference, and 95\% intervals from 10,000 paired task-level bootstrap resamples; all repetitions of a sampled task travel as one block.  Tasks are the resampling unit.

\FloatBarrier
\section{Results}
\label{sec:results}

\subsection{Controlled comparison of coordination interfaces}

Table~\ref{tab:abc-results} reports MetaSkill relative to each baseline.  Quality-preference point estimates exceed parity on both backbones. Against \methodB{} under DeepSeek-V4-Flash, MetaSkill has a preference estimate of \DeepCBPref{} (95\% task-bootstrap CI \DeepCBCI{}).

\begin{table}[!htbp]
\centering
\caption{Coordination-interface comparison: blinded pairwise artifact preference.}
\label{tab:abc-results}
\begin{minipage}{\linewidth}
\papertablestyle
\begin{tabular*}{\linewidth}{@{\extracolsep{\fill}}l c c c c@{}}
\toprule
\textbf{Backbone} & \textbf{Baseline} & \textbf{W/T/L} & \textbf{Preference} $\uparrow$ & \textbf{95\% task CI} \\
\midrule
GLM-5.2 & \methodA{} & \GLMReviewedCAWTL{} & \GLMReviewedCAPref{} & \GLMReviewedCACI{} \\
        & \methodB{} & \GLMReviewedCBWTL{} & \GLMReviewedCBPref{} & \GLMReviewedCBCI{} \\
\addlinespace[4pt]
DeepSeek-V4-Flash & \methodA{} & \DeepCAWTL{} & \DeepCAPref{} & \DeepCACI{} \\
                   & \methodB{} & \DeepCBWTL{} & \DeepCBPref{} & \DeepCBCI{} \\
\bottomrule
\end{tabular*}
\papertablenote{All outcomes are from MetaSkill's perspective. Each row contains 60 pairs (20 tasks $\times$ 3 repetitions); 95\% CIs use the task-blocked bootstrap. }
\end{minipage}
\end{table}

The GLM-5.2 trace audit shows complete node execution and declared precedence conformance: \RuntimeCompletedNodes{}/\RuntimeExpectedNodes{} nodes complete, there are \ConformanceDependency{} declared dependency/precedence violations, and \ConformanceBinding{} declared predecessor-text references identify recorded nodes and have non-empty rendered input values.  In addition, \ConformanceParallel{} traces contain overlapping execution of ready nodes.  These observations show that MetaSkill makes the intended coordination persistent and auditable.

\FloatBarrier
\subsection{MetaSkill versus AgentSkillOS with matched Skills}

In Table~\ref{tab:agentskillos-results}, the artifact-preference point estimate favors MetaSkill (\AgentMatchedPref{}, 95\% task-bootstrap CI \AgentMatchedCI{}) under matched Skill availability.

\begin{table}[!htbp]
\centering
\caption{MetaSkill versus AgentSkillOS with matched Skills: blinded pairwise artifact preference.}
\label{tab:agentskillos-results}
\begin{minipage}{\linewidth}
\papertablestyle
\begin{tabular*}{\linewidth}{@{\extracolsep{\fill}}l c c c@{}}
\toprule
\textbf{Setting} & \textbf{W/T/L} & \textbf{Preference} $\uparrow$ & \textbf{95\% task CI} \\
\midrule
Matched Skills & \AgentMatchedWTL{} & \AgentMatchedPref{} & \AgentMatchedCI{} \\
\bottomrule
\end{tabular*}
\papertablenote{Outcomes are from MetaSkill's perspective across 60 pairs (20 tasks $\times$ 3 repetitions), using the same MetaSkill artifact set. Both systems receive the same Skill packages; retrieval is outside the comparison.}
\end{minipage}
\end{table}

\FloatBarrier
\subsection{MetaSkill versus agent stacks}

All three agent-stack preference point estimates favor MetaSkill in Table~\ref{tab:harness-results}; the OpenClaw interval excludes parity (\NativeFPref{}, \NativeFCI{}).  The OpenSquilla and Hermes intervals include parity.

\begin{table}[!htbp]
\centering
\caption{Agent-stack comparison under GLM-5.2: blinded pairwise artifact preference.}
\label{tab:harness-results}
\begin{minipage}{\linewidth}
\papertablestyle
\begin{tabularx}{\linewidth}{@{}Y c c c@{}}
\toprule
\textbf{Comparator} & \textbf{W/T/L} & \textbf{Preference} $\uparrow$ & \textbf{95\% task CI} \\
\midrule
OpenSquilla (NL-SOP) & \NativeAWTL{} & \NativeAPref{} & \NativeACI{} \\
OpenClaw & \NativeFWTL{} & \NativeFPref{} & \NativeFCI{} \\
Hermes Agent & \NativeGWTL{} & \NativeGPref{} & \NativeGCI{} \\
\bottomrule
\end{tabularx}
\papertablenote{All outcomes are from MetaSkill's perspective; each row contains 60 pairs (20 tasks $\times$ 3 repetitions). The MetaSkill artifact set is shared with the GLM-5.2 rows in Table~\ref{tab:abc-results}. The OpenSquilla (NL-SOP) row reports the same 60 artifact pairs as the GLM-5.2 NL-SOP comparison in Table~\ref{tab:abc-results}.}
\end{minipage}
\end{table}

\FloatBarrier
\section{Analysis and implications}

\paragraph{Executable coordination.}
MetaSkill turns declared coordination into runtime operations, separating the enforcement of invocation order and text handoffs from the model's local task judgment. In the audited GLM-5.2 runs, all planned nodes completed without declared precedence violations, and recorded argument fields made predecessor-text references inspectable. The execution audit and blinded artifact judgments therefore provide complementary evidence about recorded coordination and relative artifact quality.

\paragraph{Composition around complete Skills.}
Complete Skill packages can be reused across compositions, with objectives, dependencies, and predecessor-text bindings assigned per invocation. Changes to cross-Skill relationships can therefore be expressed in the composition while each package retains its instructions, scripts, references, and assets. This provides a common coordination interface for heterogeneous artifact-producing Skills without translating their local procedures into a workflow-specific language.

\section{Limitations}
\label{sec:limitations}

Our study focuses on Skill composition after the relevant component Skills have been identified, and therefore does not systematically evaluate Skill retrieval from large or evolving Skill libraries. Future work will study how retrieval quality interacts with MetaSkill construction and execution at ecosystem scale.

Artifact quality is evaluated by a single fixed judge using text and rendered visual evidence under input limits. The resulting preferences may depend on the judge and the artifact evidence available to it.

\section{Conclusion}

MetaSkill gives invocation dependencies and predecessor-text bindings executable semantics while retaining complete Skill packages and their local natural-language instructions.
Across \NumTasks{} tasks, artifact-preference point estimates favor MetaSkill in all reported comparisons, with 95\% task-bootstrap intervals excluding parity against \methodB{} under DeepSeek-V4-Flash and OpenClaw under GLM-5.2.
In audited GLM-5.2 traces, all planned nodes complete and all declared precedence constraints are satisfied.
These findings support selective formalization at Skill boundaries as a practical design for composing open-ended capabilities: local task judgments remain within Skills, while cross-invocation coordination becomes explicit, executable, and inspectable.

\end{document}